\documentclass[conference]{IEEEtran}
\IEEEoverridecommandlockouts
\usepackage{cite}
\usepackage{amsmath,amssymb,amsfonts}
\usepackage {tikz}
\usetikzlibrary {positioning}
\usepackage{multirow}
\usepackage{caption}
\usepackage{diagbox}
\usepackage{graphicx}
\usepackage{textcomp}
\usepackage{xcolor}
\usepackage{calrsfs}
\usepackage{algorithm}
\usepackage{algpseudocode}
\usepackage{array}
\newcolumntype{C}[1]{>{\centering\arraybackslash}m{#1}}
\usepackage[bookmarks=false]{hyperref}

\usepackage{hyperref}

\DeclareMathAlphabet{\pazocal}{OMS}{zplm}{m}{n}
\newcommand{\La}{\mathcal{L}}

\def\BibTeX{{\rm B\kern-.05em{\sc i\kern-.025em b}\kern-.08em
    T\kern-.1667em\lower.7ex\hbox{E}\kern-.125emX}}

\makeatletter

\def\ps@IEEEtitlepagestyle{%
  \def\@oddfoot{\mycopyrightnotice}%
  \def\@evenfoot{}%
}
\def\mycopyrightnotice{%
  {\footnotesize 978-1-7281-6251-5/20/\$31.00 ©2020 IEEE\hfill}
  \gdef\mycopyrightnotice{}
}    
    
\begin{document}

\title{On Variational Inference for User Modeling in Attribute-Driven Collaborative Filtering\\
}
\makeatletter
\newcommand{\linebreakand}{%
  \end{@IEEEauthorhalign}
  \hfill\mbox{}\par
  \mbox{}\hfill\begin{@IEEEauthorhalign}
}
\makeatother

\author{
  \IEEEauthorblockN{Venugopal Mani* \thanks{* Both authors contributed equally to this work.}}
  \IEEEauthorblockA{\textit{Walmart Labs} \\
    \textit{Sunnyvale, CA, USA}\\
    venugopal.mani@walmartlabs.com}
  \and
  \IEEEauthorblockN{Ramasubramanian Balasubramanian*  \footnotemark}
  \IEEEauthorblockA{\textit{Walmart Labs} \\
    \textit{Sunnyvale, CA, USA}\\
    r.balasubramanian@walmartlabs.com}
  \linebreakand 
  \IEEEauthorblockN{Sushant Kumar}
  \IEEEauthorblockA{\textit{Walmart Labs} \\
    \textit{Sunnyvale, CA, USA}\\
    skumar4@walmartlabs.com}
  \and
  \IEEEauthorblockN{Abhinav Mathur}
  \IEEEauthorblockA{\textit{Walmart Labs} \\
    \textit{Sunnyvale, CA, USA}\\
    amathur1@walmartlabs.com}
  \and
  \IEEEauthorblockN{Kannan Achan}
  \IEEEauthorblockA{\textit{Walmart Labs} \\
    \textit{Sunnyvale, CA, USA}\\
    kachan@walmartlabs.com}
}





\maketitle


\begin{abstract}
Recommender Systems have become an integral part of online e-Commerce platforms, driving customer engagement and revenue. Most popular recommender systems attempt to learn from users' past engagement data to understand behavioral traits of users and use that to predict future behavior. In this work, we present an approach to use causal inference to learn  user-attribute affinities through temporal contexts. We formulate this objective as a Probabilistic Machine Learning problem and apply a  variational inference based method to estimate the model parameters. We demonstrate the performance of the proposed method on the next attribute prediction task on two real world datasets and show that it outperforms standard baseline methods. 

\end{abstract}


\section{Introduction}\par 
\vspace{1.5 mm}
Recommender Systems have traditionally been studied from the lens of attempting to increase customer engagement by user modeling from past interactions. These interactions are often collected in terms of explicit user signals like ratings and item reviews. Recently, there has been a shift in literature towards building recommenders by using implicit user signals like item views, item purchases, etc. Implicit signals, while useful in increasing the coverage of user signals over items, can suffer from lack of definition of what constitutes a negative signal. This has led to a class of problems known as One Class Collaborative Filtering \cite{pan_occf_2008} where techniques like low rank approximation and negative sampling are used to improve user understanding by eliminating the ambiguity over the negative train samples.\par 
\vspace{2mm}

A common assumption in Implicit OCCF is that all positive signals are equal. However, this assumption can fail to capture the wide ranging spectrum of user interactions in some domains. Normalization techniques do exist to scale these but there exists scope for more nuanced values for the positive samples. With modern data collection capabilities, \emph{domain specific fine tuning} of user interactions can be achieved to further our understanding of abstract concepts about users (like loyalty, satisfaction with the product, etc). One such idea was introduced in \cite{lavee_satis_2019} where the concept of \emph{long term customer satisfaction} was defined through a function to track the continuous implicit signal of the user. \par 
\vspace{2mm}

While Implicit OCCF systems are quite effective, an often cited drawback of these systems for user modeling is their lack of sensitivity to temporally changing user behavior. The traits of a user from a few months prior need not necessarily model their present behavior. It follows logically to try to encode the temporal aspect of implicit signals into a user understanding objective and optimize over it. This is particularly relevant in the subset of attribute-driven collaborative filtering as users tend to develop repeat patterns on certain attributes of an item. For example, in the domain of music recommendation, affinity to artists has been studied, and the recommendation of artists similar to the ones the user is loyal to has also seen improvements in results \cite{spotify_lin_2014}. In this work, we try to further our understanding of users through the notion of \emph{temporal loyalty} and integrate it into the attribute-driven collaborative filtering framework. We optimize the objective from two sources : the transaction matrix which is a binary matrix that indicates past interaction (or lack of thereof) of the involved user with the item's attribute as well as a temporal loyalty matrix which attempts to capture drifting user loyalty over time.\par 
\vspace{2mm}

The contributions of this work are as follows: first, we model temporal loyalty of the users to augment the transaction matrix. Then, we demonstrate that optimization using variational inference over these matrices outperforms plain collaborative filtering based methods on the next attribute prediction task, thus leading to a better understanding of user preferences. The rest of the work is organized as follows: Section \ref{sec2} delves into the literature of related work, Section \ref{sec3} describes our proposed system model, Section \ref{sec4} describes our experiments on two real world datasets , Section \ref{sec5} analyzes the results, and Section \ref{sec6} concludes the work and describes possible future directions.\par 
\vspace{2mm}
\section{Related Work}\label{sec2}\par 
\vspace{2mm}

 
The idea of optimizing over two matrices for modeling user preferences is relatively new. There is active research around the kind of domain-specific objectives to be optimized for and the corresponding data that could be augmented. The authors of \cite{lavee_satis_2019} consider measures of satisfaction with the purchased items, such as the amount of time spent playing a game or the number of times a particular artist was heard. Other works focus on tasks like using dwell time in session-based recommendations \cite{bogina_dwell_2017, yi_dwell_2014} or to enrich the user-item matrix\cite{yin_dwell_2013}, leveraging implicit signals such as internet browsing logs \cite{ronen_internet_2016}, etc. On the other hand, several works exist that leverage the binary transaction matrix to tackle the well-known top-k recommendation problem in large-scale datasets, such as those dealing with memory-based collaborative filtering for explicit feedback \cite{aiolli_topn_2013}, item-based collaborative filtering to address scalability concerns\cite{deshpande_item_2004}, using stratified SGD to deal with large-scale matrix factorization \cite{gemulla_dsgd_2011}, etc. However, these rely solely on explicit user signals and fail to incorporate any temporal signals.\par 
\vspace{2mm} 

In our work, we use the temporal loyalty to an item attribute as the second matrix, thus leveraging both explicit and temporal signals to set up an optimization over the two matrices. The use of loyalty is motivated by works such as \cite{bhagat_buy_2018}, where the authors model consumers' repeat purchase behavior, as well as our experience in the domain of e-Commerce and grocery. Attribute-based collaborative filtering has been explored before in works such as \cite{zhao_attr_2018} where the authors use categorical attributes to improve recommendation through multi-task learning or hierarchical classification, and \cite{chen_attr_2020} which deals with attribute-aware collaborative filtering. Our work captures the changing affinity of the users to these attributes, and thus could be used as a first stage in hierarchical classification algorithms: to predict which brands the users will buy next, before recommending particular items of that brand.\par 
\vspace{2mm}

In terms of the application of variational inference and Bayesian statistics to solve collaborative filtering problems, most works focus on the use of Variational Auto Encoders. For example,  \cite{liang_vae_2018} introduces a generative model with a multinomial likelihood and uses Bayesian inference for parameter estimation, \cite{zheng_vae_2020} uses VAE to alleviate the problem of poor robustness and over-fitting caused by large-scale data, etc. Other works using Bayesian inference, such as \cite{kim_bayesian_2014}, which presents a scalable inference for Variational Bayesian matrix factorization with side information, or \cite{chen_vae_2018}, which proposes a distributed memo-free variational inference method for large-scale matrix factorization problems, address some of the well-known shortcomings of the same in recommender systems.\par 
\vspace{2mm} 

The experimental framework that we have adopted is called the Box's Loop \cite{blei_box_2014}, which is used to uncover patterns from the conditional distribution of a latent variable model and use them to model the data and make predictions. This was well suited for our problem, since we assume a particular structure of the latent variables and use that to explain the data, and then use those variables for future recommendations. Related works in the sub-field of latent variable modeling for recommender systems include works such as \cite{song_regression_2016}, which uses blind regression to complete the partial user-item interaction matrix and uses the features of the users and items as the latent variables, and 
\cite{harvey_bayesian_2011}, which presents a Bayesian latent variable model for rating prediction that models ratings over each user's latent interests and each item's latent topics. Embedding based approaches to model the users and items have also been tried, in works such as \cite{he_neural_2017}, which replaces the inner product between user and item latent features used in classic matrix factorization by a neural architecture, and \cite{chen_ncf_2019}, which couples deep feature learning and deep interaction modeling with a rating matrix to improve recommendation performance. Our work fits the optimization task into this framework and generates user and item attribute embeddings that explain the data well by applying variational inference, and these user representations thus obtained help us develop a better understanding of the user preferences.\par 
\vspace{1.25mm}
\section{System Model}\label{sec3}\par 
\vspace{2mm}

\subsection{The Top-k Attribute Recommendation Problem}\par 
\vspace{2mm}
The classic top-k recommendation problem can be defined as follows : given a catalog of items $C$ containing items ${i_1, i_2, \cdots , i_n}$ and an item $a \in C$, henceforth referred to as the \emph{anchor item}, finding a ranked list of distinct items ${i_1, i_2, \cdots ,i_k} \in C$ to be offered alongside the anchor item, such that the user of an e-Commerce platform is most likely to engage with them. This engagement of users can be defined by a variety of metrics (in our case, the future purchase of the item). \par 
\vspace{2mm}

A sub-problem of the top-k recommendation problem is the attribute recommendation problem. Rather than the recommendation of the items to cause the next most likely engagement, the task is to predict on the attribute. For instance, in the space of e-Commerce, the attribute of interest could be the brand of the item. Given a user $u$, the task would be to predict the $k$ most likely brands that they're likely to engage with. Attribute recommendation is a slightly more well defined space as, particularly with certain attributes like brands, users are very likely to develop behaviors like loyalty towards certain brands. While in the item space, repurchasing of an item is a commonly seen behavior, especially in domains like grocery, the re-engagement with attributes is a much more commonly exhibited behavior in a wide variety of recommender system domains (songs by the same artist, books by the same author, movies starring the same actor) and solving it can hence involve harvesting this richer behavioral signal.\par 
\vspace{2mm} 

\subsection{The Temporal Loyalty Matrix}\par 
\vspace{2mm}
Most recommender systems dealing with grocery data focus on the \emph{binary transaction matrix}. Given a set of users ${u_1, u_2, \cdots , u_m}$ and a set of items ${i_1, i_2, \cdots , i_n}$, the $(p,q)^{th}$ element of the transaction matrix is 1 if $u_p$ has purchased $i_q$ within the training window, and 0 otherwise. Oftentimes, user-based collaborative filtering algorithms that use such matrices fail to pick up important signals such as a user's loyalty to a brand, how their behavior changes dynamically with time, etc. In this work, we try to capture those signals by optimizing over a temporal loyalty matrix, L, in addition to the binary transaction matrix, T.  Since we are dealing with an attribute recommendation problem, consider a set of users ${u_1, u_2, \cdots , u_m}$ and an associated set of attribute values ${v_1, v_2, \cdots , v_n}$ for a particular attribute (say, brand).  The $(p,q)^{th}$ element of the transaction matrix,\par 
\vspace{2mm}
\begin{equation}
\hspace{-20mm} T_{pq} = \begin{cases} 
1, & \parbox{5cm}{$u_p$ has bought an item with attribute value $v_q$
      at least once in the training window} \\
      \\
0, & \text{otherwise}
\end{cases}
\label{eq1}
\end{equation}
 The $(p,q)^{th}$ element of the temporal loyalty matrix is the time-decayed sum of all the purchases of attribute value $v_q$ made by user $u_p$, that is,

\begin{equation}
\hspace{-10mm} L_{pq} = \begin{cases}
$$\displaystyle \sum_{t=t_1}^{t_k} 2^\frac{t-t_{start}}{t_{end}-t_{start}}$$, &
\parbox{4cm}{$u_p$ has bought an item with attribute value $v_q$
      $k (\geq 1)$ times in the training window} \\
      \\
0, & \text{otherwise}
\end{cases}
\label{eq2}
\end{equation}
\par
\vspace{2mm}
The variables $t_{start}$ and $t_{end}$ in Equation \ref{eq2} represent the start and end times of the training window, and $t_1, t_2,...,t_k$ are the time instances when the user purchased items with that attribute value. For instance, a particular brand of beer purchased over a year ago should not get the same weight as the one purchased a week ago as user preferences might have changed. We further show that this framework works well for recommender systems that have some notion of loyalty/preference, such as readers' predilection for certain authors, etc. This optimization \cite{lavee_satis_2019} balances data from both the transaction matrix and the temporal loyalty matrix.\par 
\vspace{2mm}

\subsection{The Bayesian Framework}\par 
\vspace{2mm}
Probabilistic Machine Learning (PML) is a sub-field of Machine Learning where domain knowledge and assumptions about the hidden structure of data are leveraged to explain the observed data. PML models large, interesting, and  interconnected datasets at scale.\par 
\vspace{2mm}
The iterative probabilistic pipeline, coined \emph{Box's Loop} by \cite{blei_box_2014} 
lists the steps of modeling a PML pipeline as \emph{positing a model} with assumptions about the hidden structure of data, \emph{inferring} the hidden variables, and \emph{criticizing} the model (the evaluation step). If the evaluation does not meet the standard required, the values of the hidden variables are revised to better explain the data at hand.\par 
\vspace{2mm}
The structure of collaborative filtering by Matrix Factorization effectively lends itself to Box's Loop. The entries of the transaction matrix and the temporal loyalty matrix constitute the observed data. The classic matrix factorization problem involves decomposition of a given Matrix $M$ into latent factor matrices $U$ and $V$ along with their respective biases $B_u$ and $B_v$. The matrices $U$ and $V$ are known as \emph{embedding matrices} in the setting of collaborative filtering. The learning task then becomes to learn the embeddings and biases such that their probability, given the observed transaction and temporal loyalty matrices is maximized. This is also known as the \emph{posterior distribution}. \par 
\vspace{2mm}

\subsection{The Posterior Distribution}\par 
\vspace{2mm}

We chose to model both the matrices, T and L, as well as the priors with normal distributions. This is logical because the values in the L matrix are continuous and distributions from the exponential family have shown good results in literature \cite{lavee_satis_2019}. Also,  the normal family of distributions is conjugate to itself (or self-conjugate) with respect to a normal likelihood function, and conjugacy has desirable properties, such as yielding a closed-form expression for the posterior.\par 
\vspace{2mm}

Fig. \ref{fig:graphical_model} is a probabilistic graphical representation of our latent variable model. It shows how the random variables depend on each other in our generative process. Thus, it helps us form the posterior by connecting the assumptions that we made about the data to the model. The components of this graphical model are the ones used in standard graphical models in the field of machine learning, such as \cite{blei_box_2014}: the nodes represent random variables, the edges represent a dependence between the nodes that they connect, and the plates denote replication. Each entry in the transaction matrix, $T_{pq}$, and the temporal loyalty matrix, $L_{pq}$, depends only on its local variables $u_p, bu_p, v_q,$ and $bv_q$, which are the embedding and the bias vectors of the $p^{th}$ user, and the embedding and the bias vectors of the $q^{th}$ attribute respectively, and the corresponding global variables ($\kappa$ and $\psi$), as is the case with conditionally conjugate models. $\kappa$ and $\psi$ are the scale and the location parameters, which allow the distributions of T and L to have different dynamic ranges and be centered around different means, despite sharing some parameters which model the positive correlation between the transactions and the temporal loyalty scores, as seen in works like \cite{lavee_satis_2019}. \par
\vspace{2mm}
As mentioned earlier, all these variables have normal priors with mean 0 (except the scale parameters, which have a mean of 1) and a variance that depends on a hyperparameter, denoted by $\alpha$ with a subscript corresponding to the variable, as seen in equation \ref{eq3}.  Utilising a modeling strategy similar to \cite{lavee_satis_2019}, we write out the posterior in equation \ref{eq3} as being proportional to the product of the likelihoods and the priors. H is the set of all hyperparameters: those represented by solid black circles in Figure \ref{fig:graphical_model}, $\gamma$, and $\beta$. $\gamma$ allows us to control how much importance we give to the two likelihoods relative to each other, and $\beta$ is used to model the variance.  The matrices T and L constitute the observed data, represented by grey circles in Fig. \ref{fig:graphical_model}. $\theta$ is the set of latent variables: the user and item embeddings and biases, and the location and scale parameters, represented by white circles in Fig. \ref{fig:graphical_model}. Each user and item vector is of dimension d. \par 
\vspace{2mm}

\begin{equation}\label{eq3}
\begin{split}
P(\theta | T,& L,H) \propto P(T,L|\theta,H)P(\theta|H) \\
&\hspace{-11mm}= \prod_{(p,q,T_{pq})\in T}  P(T_{pq}|u_p, v_q, bu_p, bv_q, \kappa_t, \psi_t, H) \\  &\hspace{-7mm}  \prod_{(p,q,L_{pq})\in L}  P(L_{pq}|u_p, v_q, bu_p, bv_q, \kappa_l, \psi_l, H) \\
&\hspace{-2mm}  \prod_{p = 1}^{m} [P(u_p|\alpha_u) P(bu_p|\alpha_{bu})]
\prod_{q = 1}^{n} [P(v_q|\alpha_v) P(bv_q|\alpha_{bv})] \\
&\hspace{-0.75mm} P(\kappa_t|\alpha_{\kappa_t}) P(\psi_t|\alpha_{\psi_t}) P(\kappa_l|\alpha_{\kappa_l}) P(\psi_t|\alpha_{\psi_l})\\
&\hspace{-11mm}= \prod_{(p,q,T_{pq})\in T} \mathcal{N}(\kappa_t(u_p^{T}v_q+bu_p+bv_q)+\psi_t, (\gamma \beta)^{-1})\\
&\hspace{-7mm} \prod_{(p,q,L_{pq})\in L} \mathcal{N}(\kappa_l(u_p^{T}v_q+bu_p+bv_q)+\psi_l, ((1-\gamma) \beta)^{-1})\\
&\hspace{-2mm} \prod_{p = 1}^{m} [\mathcal{N}(0,\alpha_u^{-1} \boldsymbol{I_d}) \mathcal{N}(0,\alpha_{bu}^{-1})] \\
& \hspace{-2mm}\prod_{q = 1}^{n} [\mathcal{N}(0,\alpha_v^{-1} \boldsymbol{I_d}) \mathcal{N}(0,\alpha_{bv}^{-1})]\\
& \hspace{-1mm}\mathcal{N}(1, \alpha_{\kappa_t}^{-1}) \mathcal{N}(0, \alpha_{\psi_t}^{-1}) \mathcal{N}(1, \alpha_{\kappa_l}^{-1}) \mathcal{N}(0, \alpha_{\psi_l}^{-1})
\end{split}
\end{equation}

\subsection{Variational Inference}\par 
\vspace{2mm}
The posterior  $P(\theta | T, L,H)$ from Equation \ref{eq3} solves for the family of high dimensional latent variables  $\theta$, given the initial prior distributions over the latent variables and a likelihood function $P(T,L| \theta, H)$ that we posit about the model. The direct application of Bayes' theorem has the problem of an intractable high dimensional integration in the denominator and hence an approximate Bayesian inference of the posterior is carried out. One of the most popular methods for approximate Bayesian inference is the Markov Chain Monte Carlo (MCMC). However, despite the high accuracy of MCMC, the scale of our problem means that it is virtually impossible to use due to the computational time involved. \par 
\vspace{2mm}

Instead, a scalable approach is to treat the posterior approximation as an optimization problem through variational inference \cite{blei_ptm_2012, blei_review_2017}. The objective of variational inference is to find the \emph{variational distribution} which is a proxy-posterior $q$ parametrized by $\nu$, such that the variational distribution is least-divergent from the true posterior $p$. We adopted the widely used Kullback–Leibler divergence (KL Divergence) as the divergence metric between the two distributions. The KL divergence term is an intractable one and the equivalent of minimizing the KL-Divergence is the maximization of the Evidence Lower Bound (ELBO) \cite{braun_elbo_2010}. The ELBO, $\La(\nu)$, is described in equation \ref{eq_elbo}.\par 
\vspace{1mm}
\begin{equation}
\label{eq_elbo}
 \La(\nu) = \mathbb{E}_{q}[log(p(T,L|\theta))] - KL(q(\theta;\nu) \parallel p(\theta))
\end{equation}\par 
\vspace{1mm}

The terms here provide the classical Bayesian trade off between the log likelihood of the data and the prior over the parameters of the model. That is, the first term tries to maximize the likelihood of the observed transactions and the temporal loyalty scores, given the embedding vectors. The second term is the KL divergence between the the variational distribution and the prior over the embedding vectors. The second term effectively acts as a regularizer as it tries to minimize the divergence from the prior and hence prevents the optimizer from converging to the maximum likelihood estimate.\par 
\vspace{2mm}

Stochastic gradient descent is the commonly used approach to optimize the ELBO objective. Some works \cite{lavee_satis_2019} also use coordinate descent and other variants of gradient descent to compute the gradients and update the parameters.  The gradients can be obtained by rewriting equation \ref{eq_elbo} in terms of the complete log likelihood and then computing the gradient, as shown in equation \ref{eq5}.
\begin{equation} 
\label{eq5}
\nabla_\nu \La(\nu) = \nabla_\nu \mathbb{E}_{q}[log(p(T,L,\theta)) - log(q(\theta;\nu))]
\end{equation}\par 
\vspace{1mm}

In this work, we use score function gradient estimators \cite{ranganath_bb_2014}, \cite{paisley_variational_2012}, by rewriting equation \ref{eq5} as 
\begin{equation} 
\label{eq6}
\nabla_\nu \La(\nu) = \mathbb{E}_{q}[\nabla_\nu log(q(\theta;\nu)) (log(p(T,L,\theta)) - log(q(\theta;\nu)))]
\end{equation}\par 
\vspace{2mm}

\subsection{Prediction function}\par 
\vspace{2mm}
Once the variational distribution is approximated, predictions are made from the \emph{posterior predictive function}. The posterior predictive function uses the likelihood function from Equation \ref{eq3},  $P(T,L|\theta,H)$ to generate the predictions. After the estimation of the latent variables $\theta$, the values of the transaction entry $T_{pq}$ and temporal loyalty entry $L_{pq}$ for user $p$ and attribute $q$ are estimated from the distributions $\mathcal{N}(\kappa_t(u_p^{T}v_q+bu_p+bv_q)+\psi_t, (\gamma \beta)^{-1})$  and $\mathcal{N}(\kappa_l(u_p^{T}v_q+bu_p+bv_q)+\psi_l, ((1-\gamma) \beta)^{-1})$ respectively. Once the two values are determined, a simple addition of the two values gives the overall score for that particular user-attribute pair. \par 
\vspace{2mm}


\begin{figure}[t]
\begin{center}
\resizebox{250pt}{!}{\includegraphics{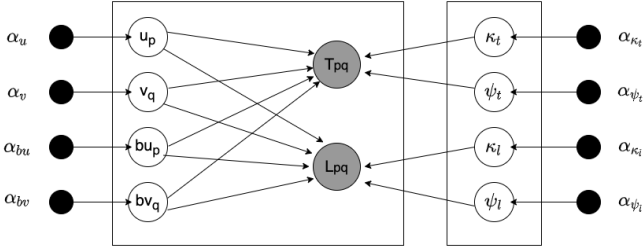}}
\caption{A graphical representation of the proposed latent variable model}
\label{fig:graphical_model}
\end{center}
\end{figure}

\section{Experiments}\label{sec4}\par 
\vspace{2mm}
\subsection{Datasets and preprocessing}\par 
\vspace{2mm}

We demonstrate our results on two datasets from different domains. The first is a private dataset from a large-scale e-Commerce company. We collected six months' worth of grocery transaction data. From the transaction metadata, the customer id, the brand, the transaction date, and the event epoch (exact epoch at which the transaction happened) were chosen. Since we are trying to understand and model loyalty, we decided to filter out customers that didn't have more than a threshold number of items in their basket, thus keeping only \emph{engaged customers} in the dataset. From experience, we have seen that there are a few large merchants and clients, such as grocery stores, that place bulk orders. These would not be representative of a single customer and hence we decided to filter out those customers that had more than an upper threshold of transactions as well. We used the first five months of data as our train set and the final month as the test set. We also filtered out customers that weren't present in both the train and the test set, since we would not have embeddings for customers we have not seen. The other problem that both the baseline models and our model suffered from was the introduction of new brands during the test period, which happens due to change in consumer demand, seasonality effects, etc. We removed the brands not present in both the train and the test sets as well. This resulted in a dataset containing 180,000 customers and 11,200 brands.\par 
\vspace{2mm}

The second dataset is the publicly available Goodreads Book Reviews dataset. This contains ratings, reviews, and a lot of other attributes of the items and the users, such as user-book interactions, metadata of the books, etc., collected in 2017 by scraping users' public shelves on Goodreads \cite{wan_monotonic_2018, wan_spoiler_2019}. The group that collected the dataset recommends using a subset (by genre) of the dataset, as the entire dataset is really large. In keeping with our theme of loyalty, we decided to go ahead with the `fantasy and paranormal' genre. Here, we are trying to assess readers' loyalty to authors. And this genre had a high density of such interactions, as was expected due to the presence of sequels, authors that write multiple books with similar themes, etc. The relevant columns in this case were author, user id, and the time when the book was marked as read. Even though we had information about the time the book was shelved, we felt that that would be a weaker (albeit denser) signal, similar to adding an item to cart in the grocery world, and hence decided to go ahead with the time the book was marked `read', which is analogous to a transaction. We again filtered the data in a fashion similar to the one described for the first dataset, and ended up with a 150,000 users and 11,400 authors. One thing to note is that the interactions in this dataset weren't as dense as the first dataset.\par 
\vspace{2mm}

\subsection{Comparison methods}\par 
\vspace{2mm}
To compare our model, we chose the standard baselines in literature \cite{wan_monotonic_2018} : Popularity Model (Pop) and classic Matrix Factorization model (MF). The popularity model captures the popularity of attributes across each customer and recommends the most popular attributes for them. \par 
\vspace{2mm}

The second model is a standard implicit OCCF Matrix Factorization. The observed transaction data is used to learn latent factors for the users and the attributes and to predict user-attribute interactions. This was done using the standard Alternating Least Squares (ALS) optimization.\par 
\vspace{2mm} 

We studied these under two settings : the first setting is a more realistic setting with the notion of \emph{explore-exploit (EE)} built into the recommender system. Most real life recommenders employ a strategy to diversify their recommendations in the hope of increasing exposure of items which do not have much user interaction. The second setting removes the explore-exploit strategy from the two baselines to give a sterner test to our model. We also included a weighting to favor attributes that the user has prior interactions with. The baseline models and our model are compared across the metrics that are described in subsection \ref{eval}.\par 
\vspace{2mm}
\subsection{Evaluation metrics} \label{eval}\par 
\vspace{2mm}

The ground truth dataset was the list of brands bought by the users in the grocery dataset or the list of authors whose books were read by the readers in the Goodreads dataset, in the test window. The predictions from the model were a list of brands/authors, ordered by the probability that the given user would buy/read the given brand/author in the test window. We compare our model with the baseline models on five different evaluation metrics, most of them well-known in the collaborative filtering literature \cite{zhu_recall_2004, tharwat_classification_2018, manning_info_2008, jarvelin_ir_2017, lioma_eval_2017}.\par 
\vspace{2mm}
\subsubsection{NDCG@k}\par 
\vspace{2mm}
As is known, DCG works on the idea that highly relevant entries appearing lower in the predictions list returned by the models should be penalized. In our case, the relevance for a brand/author is 1 if it appears in the top k predictions for a user and is present in the ground truth, and 0 otherwise. Ideal DCG (IDCG) is used to normalize this score to account for the varying lengths of the recommendation lists returned for different users. Finally, we take a mean of the NDCG values over all the queries, which are the users in the test set, to get a measure of the performance. In the following formulae, $rel_i$ denotes the relevance of the entry at the $i^{th}$ position in the predictions list returned by the models.
$$DCG_k = \sum_{i=1}^{k} \frac{rel_i}{log_2(i+1)}, NDCG_k = \frac{DCG_k}{IDCG_k}$$\par 
\vspace{2mm}

\subsubsection{MAP@k}\par 
\vspace{2mm}
The area under the precision-recall curve, which is obtained by plotting the precision and recall at every position in a ranked list of predictions, is called the average precision. Mean of the average precision scores over a set of queries i.e. users, gives the MAP. In the following formulae, AP is the average precision, MAP is the mean average precision, P(i) is the precision at position i, $\Delta$r(i) is the change in recall from position i-1 to i, \#rel is the total number of relevant brands/authors for that user (up to k), and $|U|$ is the number of users in the test set. 
$$AP = \sum_{i=1}^{k}P(i){\displaystyle \Delta} r(i) = \frac{\sum_{i=1}^{k}P(i)rel_i}{\text{\#rel}}, MAP = \frac{\sum_{j=1}^{|U|} AP_j}{|U|}$$\par 
\vspace{2mm}

\subsubsection{Hit Rate@k}\par 
\vspace{2mm}
Essentially the true positive rate, where a true positive is a brand/author predicted in the top k that is present in the ground truth for that query (user). We take a mean of the hit rate values over all the queries (users) and report that in section \ref{sec5}.
$$\text{Hit rate} = \frac{\text{Number of True Positives}}{\text{Number of Positives}}$$\par 
\vspace{2mm}

\subsubsection{MRR@k}\par 
\vspace{2mm}
The reciprocal rank of a query response, i.e. predictions for a user, is the inverse of the position of the first item in the predictions list that is present in the ground truth for that query. Here, we consider only the first k predictions, and average the reciprocal ranks over all the users. In the following formula, $pos_i$ represents the position of the first prediction for the $i^{th}$ user that is present in the ground truth list for that user. 
$$MRR = \frac{1}{|U|} \sum_{i=1}^{|U|} \frac{1}{pos_i}$$\par 
\vspace{2mm}

\subsubsection{Limited AUC@k}\par 
\vspace{2mm}

The ROC curve is a plot of the true positive rate against the false positive rate at various threshold values, and the general objective in recommender systems is to maximize the area under the ROC curve. But, in most such settings, the entries at the top of a list are more impactful than those at the bottom, but AUC is equally affected by swaps at different places in the returned list. To address this, we use limited AUC \cite{schroder_goals_2011}, which basically is the area under the part of the curve formed by the top k recommendations. This assumes that all the other relevant recommendations (apart from the top k) are distributed uniformly over the rest of the ranking list until all entries are retrieved. Thus, a straight line is drawn between the end point of the curve formed by these k recommendations and (1,1), the upper-right point of any ROC curve, and the area thus obtained is measured. This addresses some of the issues mentioned before, since swaps below the top k don't affect the AUC. This also has a few other good properties, such as a top-k list that contains more relevant entries will yield a higher AUC score, with the order mattering if the length of the list is close to the total number of brands/authors. We take a mean over all the queries (users) to get a mean LAUC.\par 
\vspace{2mm}

\subsection{Implementation details}\par 
\vspace{2mm}

To generate the baselines, we used Turicreate \cite{tc_apple}, an open source toolkit for generating core machine learning models including recommenders. For the popularity recommender, we used the popularity recommender class and for the Matrix Factorization based model, we utilized the factorization recommender class. K-Fold cross validation was performed on both classes of models using the in-built capability to tune the models and finally the best models of each were selected for comparison. \par 
\vspace{2mm}

For our model, we wrote a custom training loop and used Edward2 \cite{tran_edward_2016, tran_deep_2017} to do black-box variational inference \cite{ranganath_bb_2014}. Edward2 is a low-level language for specifying probabilistic models as programs and performing computations. We fed the models/distributions as functions whose inputs were the random variables that we were conditioning on and the outputs were the random variables that the probabilistic program was over. In the training loop, we first computed the log-likelihood using samples from the variational distribution. We used Edward's and TensorFlow's tracing functionalities (in steps 8 and 12, Algorithm \ref{vi_edw}) to record the model's computations for automatic differentiation. We then computed the KL divergence between the variational distribution and the prior distribution using the attributes of TensorFlow's distributions, and combined that with the log likelihood obtained from the posterior predictive function to get the ELBO. We tried different optimizers, learning schedules, and hyperparameter settings. A pseudocode of Edward's custom training loop adapted to our problem setting has been presented in Algorithm \ref{vi_edw}. This loop is called a certain number of times (to ensure convergence) for each batch in each epoch and the values of the variational parameters used to build the variational distribution (step 2, algorithm \ref{vi_edw}) are the updated values (step 13, algorithm \ref{vi_edw}) from the previous run.\par 
\vspace{2mm}

\begin{algorithm}
\scriptsize
  \caption{Variational Inference Training Loop}\label{vi_edw}
  \begin{flushleft}
        \textbf{INPUT:} Batch from Transaction matrix $T_b$, batch from Temporal Loyalty matrix $L_b$, Transaction matrix T, Temporal Loyalty matrix L, set of hyperparameters H, set of latent variables $\theta$, set of prior variables $\{u, v, bu, bv, \kappa_t, \psi_t, \kappa_l, \psi_l\}$
\end{flushleft}
  \begin{algorithmic}[1]
    \Procedure{Custom training loop}{$T_b$, $L_b$}
    
    \State variational\_family, trainable\_parameters $\gets$ Build variational distribution
    \State $qu, qv, qbu, qbv, q\kappa_t, q\psi_t, q\kappa_l, q\psi_l$ $\gets$ Sample posterior variables from the variational\_family
    \State ${PP}_T, {PP}_L$ $\gets$ Obtain posterior predictive functions, $P(T|\theta, H)$ and $P(L|\theta, H)$, from equation \ref{eq3} by setting prior variables to the sample posterior values
    \State $LL_{T_b}$, $LL_{L_b}$ $\gets$ Compute the log likelihood of $T_b$ and $L_b$ from ${PP}_T$ and ${PP}_L$ respectively
    \State Initialize KL $\gets$ 0
    \For {prior\_variable, variational\_variable in [(u, qu), (v, qv), (bu, qbu),
                                        (bv, qbv), ($\kappa_t$, $q\kappa_t$), ($\psi_t$, $q\psi_r$),($\kappa_s$, $q\kappa_s$), ($\psi_s$, $q\psi_s$)]}
        \State KL $\gets$ KL + KL divergence between the distributions of the variational\_variable and the prior\_variable
        \EndFor 
    \State ELBO $\gets$ Compute ELBO using KL, $LL_{T_b}$, and $LL_{L_b}$ from equation \ref{eq_elbo}
    \State Loss $\gets$ -ELBO
    \State Get the gradients using the loss and the trainable\_parameters obtained
    \State Update the parameter values

    \EndProcedure
  \end{algorithmic}
\end{algorithm}

\section{Results and Analysis}\label{sec5}\par 
\vspace{2mm}
\begin{table}[t]
\footnotesize{
\begin{tabular}{
|C{1.1cm}|C{0.5cm}|C{0.8cm}|C{0.8cm}|C{0.8cm}|C{0.8cm}|C{0.8cm}|
}\hline
\diagbox[width=5.4em, height=7.5em]{\textbf{Metric}}{\textbf{Method}} & & \textbf{\newline \newline Pop \newline + EE} & \textbf{\newline \newline MF \newline + EE} &\textbf{\newline \newline Pop} & \textbf{\newline \newline MF} & \textbf{\newline \newline VI-\newline MF} \\  
 \hline
\multirow{6}{*}{\textbf{NDCG}}&@5 &0.047 &0.054  &0.144 &0.210 &0.212\\ 
&@10 &0.031 &0.036  &0.096 &0.140 &0.141\\
&@15 &0.026 &0.031  & 0.081 &0.118 & 0.120\\
&@20 &0.025 &0.028  &0.077 & 0.112 & 0.114\\
\hline
\multirow{6}{*}{\textbf{MAP}}&@5 &0.016 &0.021  &0.049 &0.098 &0.099\\ 
&@10 &0.008 &0.011  &0.026 &0.051 &0.053\\
&@15 &0.006 &0.008  &0.020 &0.040 &0.040\\
&@20 &0.006 &0.007 & 0.018 & 0.037 & 0.038\\
\hline
\multirow{6}{*}{\textbf{HR}}&@5 &0.064 &0.064  &0.197 &0.196 &0.198\\ 
&@10 &0.033 &0.033  &0.101 &0.101 &0.102\\
&@15 &0.024 &0.025  &0.075 &0.075 &0.076\\
&@20 &0.022 &0.022  &0.067 &0.066 &0.068\\
\hline
\multirow{6}{*}{\textbf{MRR}}&@5 &0.080 &0.108  &0.246 &0.491 &0.492\\ 
&@10 &0.080 &0.108  &0.246 &0.491 &0.492\\
&@15 &0.080 &0.108  &0.246 &0.491 &0.492\\
&@20 &0.080 &0.108  &0.246 &0.491 &0.492\\
\hline
\multirow{6}{*}{\textbf{LAUC}}&@5 &0.532 &0.532  &0.598 &0.598 &0.599\\ 
&@10 &0.516 &0.516  &0.551 &0.551 &0.552\\
&@15 &0.512 &0.512  &0.539 &0.540 &0.540\\
&@20 &0.511 &0.511  &0.536 &0.537 &0.537\\
\hline

\end{tabular}
}

\captionof{table}{Comparison of evaluation metrics across models on e-Commerce grocery data}\label{ecommerce_metrics_table}
\end{table}

\begin{table}[t]
\footnotesize{
\begin{tabular}{
|C{1.1cm}|C{0.5cm}|C{0.8cm}|C{0.8cm}|C{0.8cm}|C{0.8cm}|C{0.8cm}|
}\hline
\diagbox[width=5.4em, height=7.5em]{\textbf{Metric}}{\textbf{Method}} & & \textbf{\newline \newline Pop \newline + EE} & \textbf{\newline \newline MF \newline + EE} &\textbf{\newline \newline Pop} & \textbf{\newline \newline MF} & \textbf{\newline \newline VI-\newline MF} \\ 
 \hline
\multirow{6}{*}{\textbf{NDCG}}&@5 &0.020 & 0.027 &0.047 &0.068 &0.069\\ 
&@10 & 0.014 &0.019 &0.034 &0.049 &0.051\\
&@15 & 0.012 &0.016 &0.030 &0.043 &0.044\\
&@20 & 0.010 &0.015 &0.028 &0.040 &0.041\\
\hline
\multirow{6}{*}{\textbf{MAP}}&@5 & 0.007& 0.013 &0.017 &0.033 &0.034\\ 
&@10 & 0.004 & 0.008 &0.011 &0.021 &0.022\\
&@15 & 0.003 & 0.006 &0.009 &0.018 &0.018\\
&@20 & 0.002 & 0.005 &0.008 &0.016 &0.017\\
\hline
\multirow{6}{*}{\textbf{HR}}&@5 & 0.031 &0.028 &0.070 &0.069 &0.071\\ 
&@10 & 0.018 &0.017 &0.041 &0.041 &0.042\\
&@15 & 0.013 &0.014  &0.031 &0.031 &0.033\\
&@20 & 0.011 &0.012 &0.026 &0.027 &0.028\\
\hline
\multirow{6}{*}{\textbf{MRR}}&@5 & 0.033 & 0.061 &0.075 &0.148 &0.150\\ 
&@10 & 0.034 & 0.061 &0.075 &0.148 &0.151\\
&@15 & 0.034 & 0.061 &0.075 &0.149 &0.151\\
&@20 & 0.034 & 0.061 &0.076 &0.149 &0.152\\
\hline
\multirow{6}{*}{\textbf{LAUC}}&@5 &0.514 & 0.512 &0.533 &0.533 &0.534\\ 
&@10 & 0.508 & 0.507 &0.521 &0.521 &0.522\\
&@15 & 0.506 & 0.505 &0.517 &0.517 &0.518\\
&@20 & 0.505 & 0.505 &0.516 &0.516 &0.516\\
\hline

\end{tabular}
}

\captionof{table}{Comparison of evaluation metrics across models on Goodreads data}\label{gr_metrics_table}
\end{table}

The results on the e-Commerce data and the open source Goodreads data have been presented in Table \ref{ecommerce_metrics_table} and Table \ref{gr_metrics_table} respectively. The metrics for our model are shown in the final column, titled VI-MF (Variational Inference Matrix Factorization).\par 
\vspace{2mm}

The first two baselines, with the explore-exploit strategy, suffer from trading off accuracy for diversity and hence do not perform as well as the other models. In both settings, with and without explore-exploit, MF-based models outperform the pop models because the pop models simply recommend attributes of the items that the user has bought most in the past whereas the latent factors capture user affinities well as they learn better representations from the interactions.\par 
\vspace{2mm}

Our model shows a clear 1 to 3 percent increase in all metrics across the various ranks as compared to the best performing baseline model (that is, the classic Matrix Factorization) in both the e-Commerce grocery dataset as well as the Goodreads dataset. The size and scale of the datasets mean that these gains are significant. Quantitatively, in the e-Commerce setting, for a business with tens of billions of dollars in revenue, a 1 to 3 percent increase translates to hundreds of millions of dollars. This indicates that incorporating temporal loyalty leads to a better understanding of the user preferences, thus having an effect on the prediction of user behavior and subsequently revenue.\par 
\vspace{2mm} 

Overall, the metrics on the e-Commerce grocery data are higher than the ones on the Goodreads data. This can be explained by the higher density of grocery data leading to stronger user affinities to attributes. Interestingly, the trends across the models seem to hold across the domains of grocery and `Fantasy and Paranormal' genre. In other words, the notion of brand loyalty in grocery seems similar to the notion of author loyalty in the `Fantasy and Paranormal' genre of books.\par 
\vspace{2mm}


\section{Conclusion and Future Work}\label{sec6}\par 
\vspace{2mm}

In this work, we leverage a customer's temporal loyalty to an item attribute in addition to the engagement behavior to model their preferences and subsequently tackle the top-k attribute recommendation problem. We model this as an optimization problem over two matrices and use the Box's Loop framework and variational inference to estimate the parameter values and train the user embeddings that best explain the observed explicit and temporal signals. We demonstrate the effectiveness of the user embeddings learnt by showing that the proposed approach outperforms standard baselines for this task on a private e-Commerce grocery dataset as well as the publicly available Goodreads dataset, which also supports the hypothesis that capturing a customer's temporally changing interests can lead to better recommendations. \par 
\vspace{2mm}

In terms of future directions, one could explore other ways to come up with the loyalty scores in the Temporal Loyalty matrix, L. Some works, such as \cite{he_pseudo_2018}, also focus on enriching the transaction matrix, T, to address issues that arise due to sparsity; we plan to investigate coupling those with our current approach. Another direction to explore would be to model the priors and the likelihoods with other distributions, informed by domain knowledge and the type of data one is dealing with.\par 
\vspace{2mm}

\section{Abbreviations and Acronyms }\par 
\vspace{2mm}
ALS: Alternating Least Squares

AP: Average Precision

DCG: Discounted Cumulative Gain

EE: Explore-Exploit

ELBO: Evidence Lower Bound

HR: Hit-Rate

IDCG: Ideal Discounted Cumulative Gain

KL: Kullback-Leibler Divergence

LAUC: Limited Area Under the Curve

MAP: Mean Average Precision

MF: Matrix Factorization

MCMC:  Markov  Chain  Monte  Carlo

MRR : Mean Reciprocal Rank

NDCG: Normalized Discounted Cumulative Gain

OCCF: One Class Collaborative Filtering

PML: Probabilistic Machine Learning 

ROC: Receiver Operating Characteristic 

SGD: Stochastic Gradient Descent

VAE: Variational Auto Encoder

VI : Variational Inference

\bibliography{references}
\bibliographystyle{ieeetr.bst}

\end{document}